%% file: root.tex
\documentclass[letterpaper, 10 pt, conference]{ieeeconf}  

\IEEEoverridecommandlockouts                              

\usepackage{graphics} 
\usepackage{epsfig} 
\usepackage{times} 
\usepackage{amsmath} 
\usepackage{amssymb}  
\usepackage[]{algorithm}
\usepackage[]{algorithmic}
\usepackage{comment}
\usepackage{color}
\usepackage{multirow}
\usepackage{makecell}

\input{defs}

\title{\LARGE \bf
Learning a Unified Control Policy for Safe Falling
}

\author{Visak C.V. Kumar$^{1}$, Sehoon Ha$^{2}$, C. Karen Liu$^{1}$ 
\thanks{}
\thanks{$^{1}$ Visak C.V. Kumar and C. Karen Liu is with the School of Interactive Computing,
  Georgia Institute of Technology, GA 30308.
  {\tt\small visak3@gatech.edu, karenliu@cc.gatech.edu}}%
\thanks{$^{2}$ Sehoon Ha is with Disney Research (The work was done at Georgia Institute of Technology).
        {\tt\small sehoon.ha@disneyresearch.com}}%
}

\begin{document}

\maketitle
\thispagestyle{empty}
\pagestyle{empty}

\begin{abstract}
Being able to fall safely is a necessary motor skill for humanoids performing highly dynamic tasks, such as running and jumping. We propose a new method to learn a policy that minimizes the maximal impulse during the fall. The optimization solves for both a discrete contact planning problem and a continuous optimal control problem. Once trained, the policy can compute the optimal next contacting body part (\eg left foot, right foot, or hands), contact location and timing, and the required joint actuation. We represent the policy as a mixture of actor-critic neural network, which consists of $n$ control policies and the corresponding value functions. Each pair of actor-critic is associated with one of the $n$ possible contacting body parts. During execution, the policy corresponding to the highest value function will be executed while the associated body part will be the next contact with the ground. With this mixture of actor-critic architecture, the discrete contact sequence planning is solved through the selection of the best critics while the continuous control problem is solved by the optimization of actors. We show that our policy can achieve comparable, sometimes even higher, rewards than a recursive search of the action space using dynamic programming, while enjoying 50 to 400 times of speed gain during online execution.

\end{abstract}

\section{Introduction}
The advent of reinforcement learning in recent years has expanded the capability of agents to perform highly dynamic motor tasks, such as running or jumping. However, most policy optimization methods for dynamic tasks have been demonstrated in simulation only. To deploy the recent RL algorithms on full-body humanoids in the real world, one must overcome the challenge of preventing detrimental falls during the initial exploration phase of the learning algorithm. As such, the first policy a humanoid must acquire before learning other dynamic tasks is perhaps the skill of safe fall.

As shown in Ha and Liu \cite{Ha2015}, the falling problem can be formulated as a Markov Decision Process (MDP) that solves for a contact sequence with the ground such that the damage incurred  during the fall is minimized. Their work provides an unifying view over various falling strategies designed for specific scenarios; instead of switching among a collection of individual strategies, a single unified algorithm can plan for appropriate responses to a wide variety of falls. However, using dynamic programming to solve a MDP, as proposed in \cite{Ha2015}, is computationally demanding and infeasible to deploy to the real world. Consequently, the controller can only act in an open-loop fashion without any ability to re-plan during the course of the fall. Such a controller is extremely brittle when executing a long sequence of contacts.

One promising direction toward a fast and robust falling controller is to learn a policy through reinforcement learning. Indeed, the MDP formulation of the falling problem \cite{Ha2015} makes it innately suitable for algorithms that take advantage of the recursive Bellman equation. While a rich body of literature in RL can potentially be applicable, our problem is unique in that the action space consists of both discrete and continuous variables. That is, the discrete problem of contact sequence planning and the continuous problem of joint torque optimization need to be solved simultaneously.

We proposes a new policy optimization method to learn the appropriate actions for minimizing the damage of a humanoid fall. Our algorithm learns a policy to minimize the maximal impulse occurred during the fall. The actions in this problem include discrete decisions on the sequence of body parts used to contact the ground (\eg contact the ground with left foot, right foot, and then both hands) and continuous decisions on location and timing of the contacts, as well as the joint torques of the robot. Our algorithm is based on CACLA \cite{VanHasselt2007, VanHasselt2012} which solves an actor-critic controller for continuous state and action spaces. We adapt their actor-critic model to solve for both discrete and continuous actions using the architecture, MACE, proposed by Peng \etal \cite{Peng2016}. 

Our algorithm trains $n$ control policies (actors) and the corresponding value functions (critics) in a single interconnected neural network (Figure \ref{fig_NN}). Each policy and its corresponding value function are associated with a candidate contacting body part, which is designed or preferred to be a contact point with the ground, such as hands, feet, or knees. During policy execution, the network is queried every time the robot establishes a new contact with the ground, allowing the robot to re-plan throughout the fall. Based on the current state of the robot, the actor corresponding to the critic with the highest value will be executed while the associated body part will be the next contact with the ground. With this mixture of actor-critic architecture, the discrete contact sequence planning is solved through the selection of the best critics while the continuous control problem is solved by the optimization of actors.

We demonstrate that our algorithm reliably reduces the maximal impulse induced by a fall on both simulated humanoids and on the actual hardware. Different contact strategies emerge as the initial conditions vary. We also show that the algorithm can run in real time (as opposed to 1-10 seconds reported by \cite{Ha2015}) once the policy is trained. Comparing to the actions computed by Ha and Liu \cite{Ha2015}, our policy produces better rewards on average with only $0.25\%$ to $2\%$ of computation time. 
		
\section{RELATED WORK}
\input{related}

\section{Preliminaries}
\begin{figure}
\centering
\includegraphics[width=0.9\columnwidth]{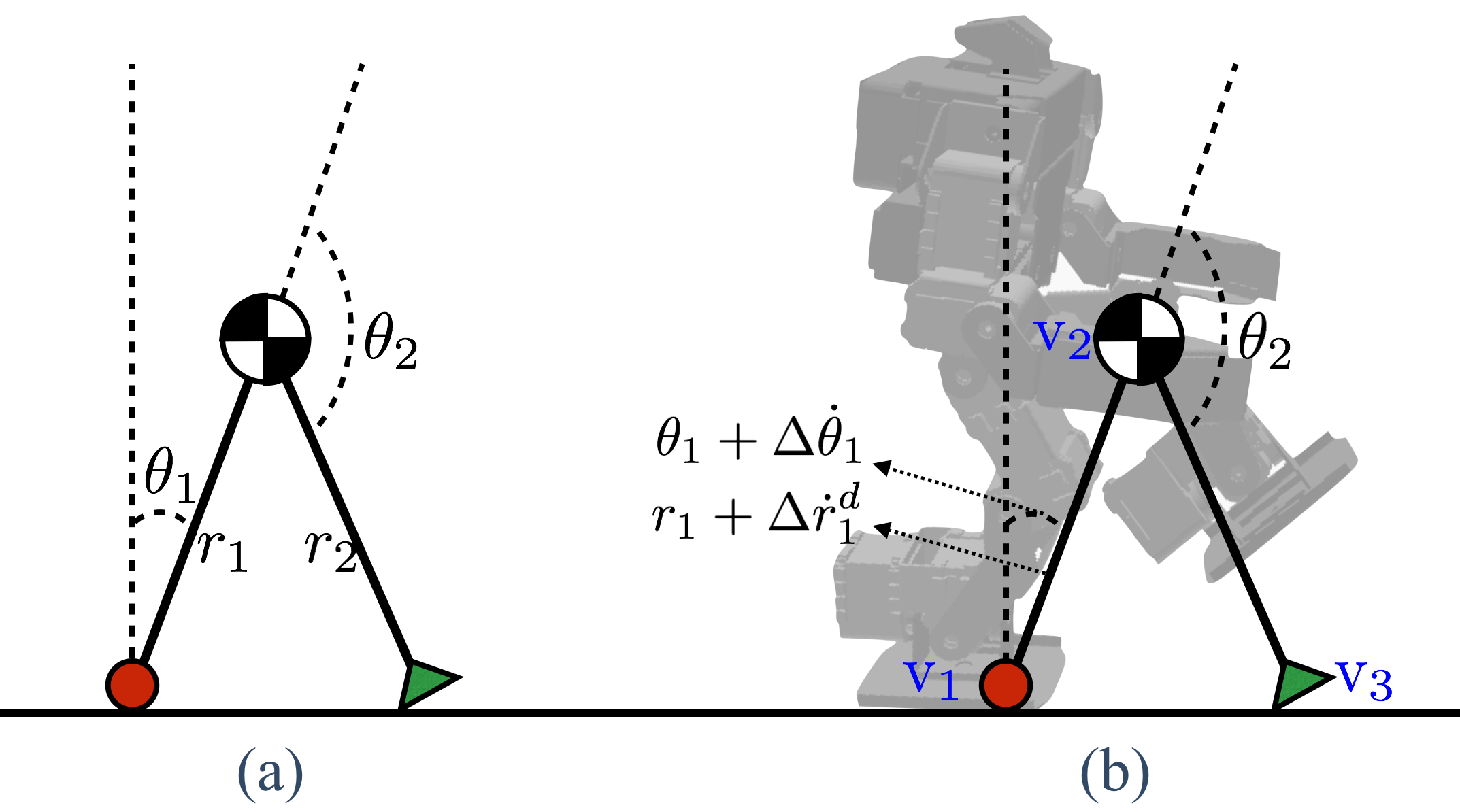}
\caption{(a) The abstract model consists of an inverted pendulum with a massless telescopic rod that connects the center of mass (COM) and the center of pressure (red circle). An additional telescopic rod, called "stopper" is also attached to the COM. $\theta_1$ and $\theta_2$ indicate the angles of two rods while $r_1$ and $r_2$ indicate their lengths. (b) The joint-level policy first predicts the COM ($\mathrm{v}_2$) and the next contact location ($\mathrm{v}_3$)  using the current state of the robot (shown as the grayed out robot figure) and the action computed by the abstract-level policy, and then solves a target full-body pose using inverse kinematics.}
\label{fig:AbsModel}
\end{figure}

For completeness we briefly review the MDP formulation of the falling problem proposed by Ha and Liu \cite{Ha2015}. The objective of the problem is to find a sequence of actions such that the maximal impulse during the fall is minimized. The value function of such an optimization can be expressed as:
\begin{equation}
  \begin{aligned}
  	V(\vc{s}) = \min_{\vc{a}} \max(g(\vc{s}, \vc{a}), V(f(\vc{s}, \vc{a}))),
  \end{aligned}
\end{equation}
where $\vc{s} \in \mathcal{S}$ and $\vc{a} \in \mathcal{A}$ are the state space and the action. The function $g(\vc{s}, \vc{a}): \mathcal{S} \times \mathcal{A} \mapsto \mathbb{R}$ evaluates the impulse induced by taking the action $\vc{a}$ at the state $\vc{s}$. The transition function, $\vc{s}' = f(\vc{s}, \vc{a})$, computes the next state $\vc{s}'$ by integrating the dynamic equations of the system.

To solve the falling problem efficiently, Ha and Liu \cite{Ha2015} proposed an abstract model to approximate the motion of the humanoid (Figure \ref{fig:AbsModel}(a)). The abstract model consists of an inverted pendulum with a massless telescopic rod that connects the center of mass (COM) and the center of pressure in the sagittal plane. Furthermore, a second massless telescopic rod, called a "stopper", is attached to the COM and used to approximate the reactive limb motion in human falls. Multiple abstract models are used to represent a sequence of contacts. As the current stopper hits the ground, a new abstract model is initialized with the pivot locating at the tip of the current stopper. This abstract model can be used to represent any contact point on the robot, but in practice the robot has only a limited number of preferred contacting body parts, such as feet or hands. As a result, there exists a finite set of contact sequences achievable by a given robot.

Using the abstract models to formulate the MDP, the state at each impact moment is defined as $\vc{s} = \{c_1, t, r_1, \theta_1, \dot{r}_1, \dot{\theta}_1\}$, where $c_1$ denotes the index of the contacting body part, $t$ denotes the elapsed time from the beginning of the fall until this impact occurs, and other parameters describe the position and the velocity of the pendulum at the impact moment detailed in \figref{AbsModel}(a). An action $\vc{a} = \{c_2, \theta_2, \Delta, \dot{r}_1^d\}$ describes the index of the contacting body part used as the stopper ($c_2$), the position of the stopper at the next impact moment ($\theta_2$), the elapse time from the previous impact moment to the next impact moment ($\Delta$), and the desired speed of the pendulum length during the current contact $\dot{r}_1^d$.

Determining the best action from a given state is a 4D search problem with a mixture of continuous and discrete action variables. Ha and Liu \cite{Ha2015} discretized the continuous action parameters and exploited the monotonic nature of the falling problem to accelerate the dynamic programming. However, the computation of the algorithm is still far from real-time ($1-10$ seconds computation time after the robot receives an external force) and impractical to deploy in the real world.

\begin{figure}[!t]
\centering
\includegraphics[width=0.8\columnwidth]{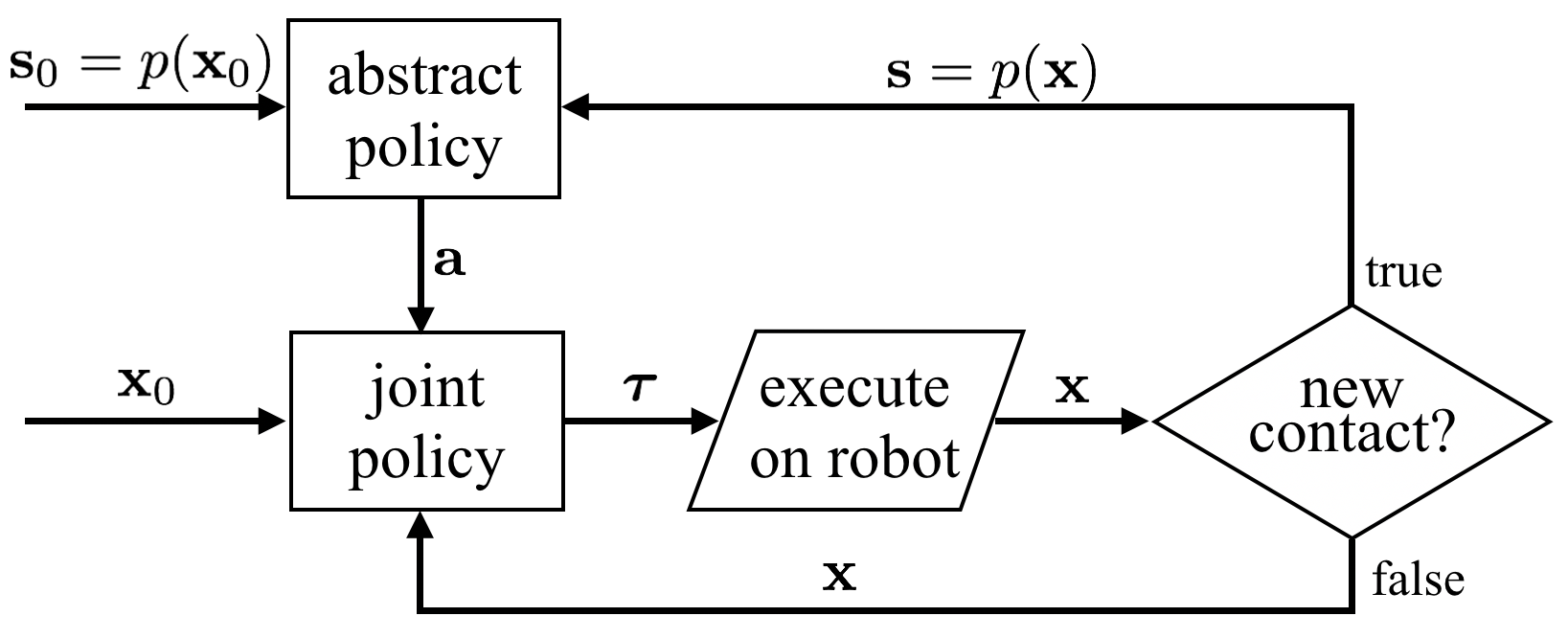}
\caption{The overall policy consists of two components: abstract-level policy and joint-level policy.}
\label{fig_ControlArch}
\end{figure}

\section{Method}
We present a new approach to the falling problem using a reinforcement learning framework. Our goal is to learn a control policy capable of computing optimal contact sequence, contact timing and locations, and joint actuation in real time. The overall policy consists of two components: abstract-level policy and joint-level policy. The abstract-level determines the optimal action on the abstract model, while the joint-level policy maps the abstract action to the full body space of the robot and produces joint actuation.

Figure \ref{fig_ControlArch} illustrates the workflow of the overall policy. We first define a function $p: \mathcal{X} \mapsto \mathcal{S}$ which maps a state of robot ($\vc{x} \in \mathcal{X}$) to a state of abstract model ($\vc{s} \in \mathcal{S}$). The mapping can be easily done because the full set of joint position and velocity contains all the necessary information to compute $\vc{s}$. As the robot receives an external force initially, the state of the robot is projected to $\mathcal{S}$ and fed into the abstract-level policy. The action ($\vc{a}$) computed by the abstract-level policy is passed into the joint-level policy to compute the corresponding joint torques ($\boldsymbol{\tau}$). If no new contact is detected after executing $\boldsymbol{\tau}$, the new state of the robot will be fed back into the joint-level policy and a new $\boldsymbol{\tau}$ will be computed (the lower feedback loop in Figure \ref{fig_ControlArch}). If the robot encounters a new contact, we re-evaluate the contact plan by querying the abstract-level policy again  (the upper feedback loop in Figure \ref{fig_ControlArch}).


\begin{figure}[!t]
\centering
\includegraphics[width=0.8\columnwidth]{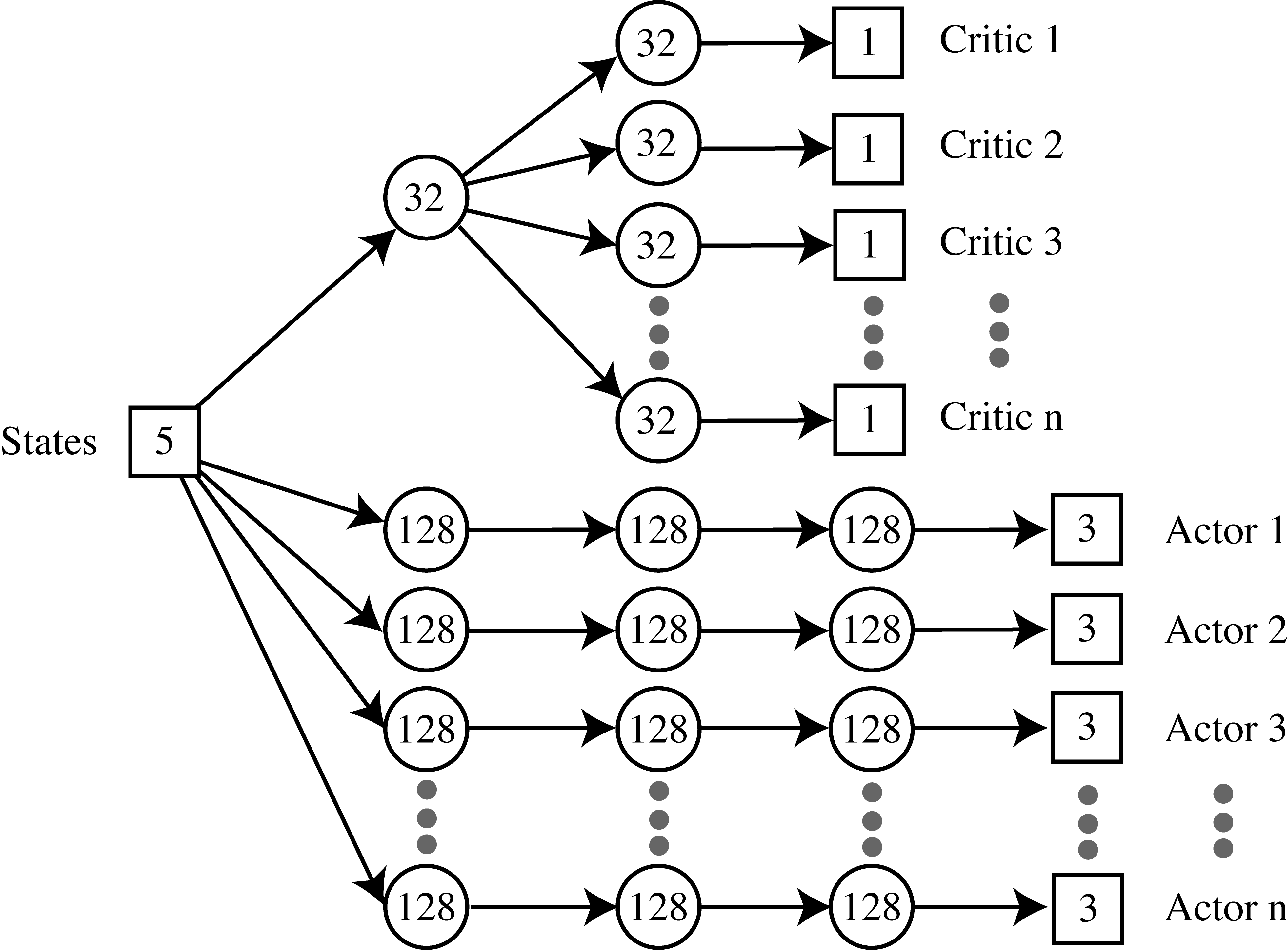}
\caption{A schematic illustration of our deep neural network that consists of $n$ actor-critics. The numbers indicate the number of neurons in each layer.}
\label{fig_NN}
\end{figure}

\subsection{Abstract-level Policy}
To overcome the challenge of optimizing over both discrete and continuous action variables, we introduce a new policy representation based on a neural network architecture inspired by MACE \cite{Peng2016}. The policy takes as input the state of the abstract model and outputs the action, as well as the next contacting body part. The state space is defined as $\vc{s} = \{c_1, r_1, \theta_1, \dot{r}_1, \dot{\theta}_1\}$, where the total elapsed time $t$ is removed from the state space previously defined by Ha and Liu \cite{Ha2015}. As a side benefit of a feedback policy, we no longer need to keep track of the total elapsed time. For the action space, we remove the discrete action variable, $c_2$ so that the new action is a continuous vector in $\mathbb{R}^3$: $\vc{a} = \{\theta_2, \Delta, \dot{r}_1^d\}$. The network combines $n$ pairs of actor-critic subnets, each of which is associated with a contacting body part. Each actor, $\Pi_i(\vc{s}): \mathbb{R}^5 \mapsto \mathbb{R}^3$, represents a control policy given that the $i$-th contacting body part is chosen to be the next contact. Each critic, $V_i(\vc{s}): \mathbb{R}^5 \mapsto \mathbb{R}$, represents the value function that evaluates the return (long-term reward) of using the $i$-th contacting body part as the next contact and taking the action computed by $\Pi_i(\vc{s})$ as the next action. We fuse all $n$ actor-critic pairs in one single network with a shared input layer (Figure \ref{fig_NN}).

At each impact moment when a new contact is established, the network evaluates all the $V_i(\vc{s}), \;1 \leq i \leq n$, and selects the policy corresponding to the highest critic. This arrangement allows us to train $n$ experts, each specializes to control the robot when a particular contacting body part is selected to be the next contact. As a result, we cast the discrete contact planning into the problem of expert selection, while simultaneously optimizing the policy in continuous space.

\subsubsection{Reward}
Since our goal is to minimize the maximal impulse, we define the reward function as:
\begin{equation}
r(\vc{s}, \vc{a}) = \frac{1}{1 + j}, 
\end{equation}
where $j$ is the impulse induced by the contact. Suppose the COM of the pendulum and the tip of the stopper are $(x_1, y_1)$ and $(x_2, y_2)$ respectively at the impact moment, the impulse can be computed as: 
\begin{equation}
j = -\frac{\dot{y}_2^-}{\frac{1}{M} + \frac{1}{I}(x_2 - x_1)^2},
\end{equation}
where $M$ is the mass and $I$ is the inertia of the abstract model (see details in \cite{Ha2015}). 

With this definition of the reward function, the objective of the optimization is to maximize the minimal reward during the fall.

\subsubsection{Learning Algorithm}
Algorithm 1 illustrates the process of learning the abstract-level policy. We represent the policy using a neural network consisting $n$ pairs of actor-critic subnets with a shared input layer (Figure \ref{fig_NN}). Each critic has two hidden layers with 32 neurons each. The first hidden layer is shared among all the critics. Each actor has 3 hidden layers with 128 neurons each. All the hidden layers use tanh as the activation functions. We define weights and biases of the network as $\theta$, which is the unknown vector we attempt to learn.

The algorithm starts off with generating an initial set of high-reward experiences, each of which is represented as a tuple: $\tau = (\vc{s}, \vc{a}, \vc{s}', r, c)$, where the parameters are the starting state, action, next state, reward, and the next contacting body part. To ensure that these tuples have high reward, we use the dynamic-programming-based algorithm described in \cite{Ha2015} (referred as DP thereafter) to simulate a large amount of rollouts from various initial states and collect tuples. Filling the training buffer with these high-reward experiences accelerates the learning process significantly. In addition, the high-reward tuples generated by DP can guide the abstract-level policy to learn "achievable actions" when executed on a full body robot. Without the guidance of these tuples, the network might learn actions that increase the return but unachievable under robot's kinematic constraints. 


In addition to the initial experiences, the learning algorithm continues to explore the action space and collect new experiences during the course of learning. At each iteration, we simulate $K (=10)$ rollouts starting at a random initial state sampled from a Gaussian distribution $\mathcal{N}_0$ and terminating when the abstract model comes to a halt. A new tuple is generated whenever the abstract model establishes a new contact with the ground. The exploration is done by stochastically selecting the critic and adding noise in the chosen actor. We follow the same Boltzmann exploration scheme as in \cite{Peng2016} to select the actor $\Pi_i(\vc{s})$ based on the probability defined by the predicted output of the critics:
\begin{equation}
\label{boltzman}
\mathcal{P}_i(\vc{s}) = \frac{e^{V_i(\vc{s}|\theta)/T_t}}{\sum_j e^{V_j(\vc{s}|\theta)/T_t}},
\end{equation}
where $T_t (=5)$ is the temperature parameter, decreasing linearly to zero in the first $250$ iterations. While the actor corresponding to the critic with the highest value is most likely to be chosen, the learning algorithm occasionally explores other actor-critic pairs, essentially trying other possible contacting body parts to be the next contact. Once an actor is selected, we add a zero-mean Gaussian noise to the output of the actor. The covariance of the Gaussian is a user-defined parameter.

After $K$ rollouts are simulated and tuples are collected, the algorithm proceeds to update the critics and actors. In critic update, a minibatch is first sampled from the training buffer with $m (=32)$ tuples, $\tau_i = (\vc{s}_i, \vc{a}_i, \vc{s}'_i, r_i, c_i)$. We use the temporal difference to update the chosen critic similar to \cite{Mnih2015}:
\begin{eqnarray}
&y_i = \min(r_i, \gamma \max_j \hat{V}_j(\vc{s}'_i)) \\ \nonumber
&\theta \leftarrow \theta + \alpha \sum_i (y_i -  V_{c_i}(\vc{s})) \nabla_\theta V_{c_i}(\vc{s})
\end{eqnarray}
where $\theta$ is updated by following the negative gradient of the loss function $\sum_i (y_i - V_{c_i}(\vc{s}))^2$ with the learning rate $\alpha (=0.0001)$. The discount factor is set to $\gamma = 0.9$. Note that we also adopt the idea of target network from \cite{Mnih2015} to compute the target, $y_i$, for the critic update. We denote the target networks as $\hat{V}(\vc{s})$.

The actor update is based on supervised learning where the policy is optimized to best match the experiences: $\min_\theta \sum_i \|\vc{a}_i - \Pi_{c_i}(\vc{s}_i)\|^2$. We use the positive temporal difference to decide whether matching a particular tuple is advantageous:
\begin{eqnarray}
&y = \max_j V_j(\vc{s}_i) \\ \nonumber
&y' = \min(r_i, \gamma \max_j \hat{V}_j(\vc{s}'_i)) \\ \nonumber
&\mathrm{if\;\;} y' > y, \;\;\;\theta \leftarrow \theta  + \alpha (\nabla_\theta \Pi_{c_i}(\vc{s}))^T(\vc{a}_i - \Pi_{c_i}(\vc{s})).
\end{eqnarray}

\begin{algorithm} 
\label{alg1}
\caption{Learning abstract-level policy}
\begin{algorithmic}[1]                    
\STATE Randomly initialize $\theta$ 
\STATE Initialize training buffer with tuples from DP 
\WHILE{not done}
\STATE EXPLORATION:
\FOR{$k=1 \cdots K$}
\STATE $\vc{s} \sim \mathcal{N}_0$ 
\WHILE{$\vc{s}.\dot{\theta}_1 \geq 0$}
\STATE c $\leftarrow$  Select actor stochastically using Equation \ref{boltzman}
\STATE \vc{a} $\leftarrow \Pi_{c}(\vc{s}) + \mathcal{N}_{t}$
\STATE Apply $\vc{a}$ and simulate until next impact moment
\STATE $\vc{s}' \leftarrow$ Current state of abstract model 
\STATE $r \leftarrow r(\vc{s}, \vc{a})$
\STATE Add tuple $\tau \leftarrow (\vc{s}, \vc{a}, \vc{s}', r, c)$ in training buffer
\STATE $\vc{s} \leftarrow \vc{s}'$
\ENDWHILE
\ENDFOR
\newline
\STATE UPDATE CRITIC: 
\STATE Sample a minibatch $m$ tuples $\{ \tau_{i} = (\vc{s}_{i}, \vc{a}_i, \vc{s}'_{i}, r_i, c_{i})\}$ 
\STATE $y_{i}$ $\leftarrow$ $\min(r_{i},\gamma \max_j$ $\hat{V}_{j}(\vc{s}'_{i}))$ for each $\tau_{i}$   
\STATE $\theta \leftarrow \theta + \alpha \sum_i (y_i -  V_{c_i}(\vc{s})) \nabla_\theta V_{c_i}(\vc{s})$
\newline
\STATE UPDATE ACTOR: 
\STATE Sample a minibatch $m$ tuples $\{ \tau_{i} = (\vc{s}_{i}, \vc{a}_i, \vc{s}'_{i}, r_i, c_{i})\}$
\STATE $y_{i} = \max_j V_j(\vc{s})$ 
\STATE $y'_{i} \leftarrow \min(r_i,\gamma \max_j \hat{V}_{j}(\vc{s}'_{i}))$ 
\IF {$y'_{i} > y_{i}$}
\STATE $\theta \leftarrow \theta  + \alpha (\nabla_\theta \Pi_{c_i}(\vc{s}))^T(\vc{a}_i - \Pi_{c_i}(\vc{s}))$

\ENDIF     
\ENDWHILE
\end{algorithmic}
\end{algorithm}

\subsection{Joint-level Policy}
The goal of the joint-level policy is to execute the action computed by the abstract-level policy in the joint space of the robot. Recall that the abstract-level policy outputs an action vector $\vc{a} = \{\theta_2, \Delta, \dot{r}_1^d\}$ and the next contacting body part $c$. Together with the current state $\vc{s} = \{c_1, r_1, \theta_1, \dot{r}_1, \dot{\theta}_1\}$, we can define a triangle formed by the pivot of the pendulum ($\mathrm{v}_1$), the mass point ($\mathrm{v}_2$), and the tip of the stopper ($\mathrm{v}_3$) (Figure \ref{fig:AbsModel}(b)). The shape of the triangle at the next impact moment can be determined by $\theta_2$, $\theta_1 + \Delta \dot{\theta}_1$, and $r_1 + \Delta \dot{r}_1^d$. The latter two terms are the predicted $\theta_1$ and the predicted $r_1$ at the next impact moment. We then transform the triangle into the coordinate frame of the robot by aligning $\mathrm{v}_1$ with the current contact point of the robot and rotating the triangle such that $\mathrm{v}_3$ touches the ground. We define the coordinates of $\mathrm{v}_2$ as the target of the robot's COM and the coordinates of $\mathrm{v}_3$ as the target of the next contacting body part, $c$. We then solve the target pose to match the two targets using inverse kinematics. Once the target pose is solved, we use a PD control to track the target pose (the lower feedback loop in Figure \ref{fig_ControlArch}) until the next impact moment occurs.

\input{results}

\input{conclusion}








\bibliographystyle{IEEEtran}
\bibliography{Falling}



\end{document}

%% file: defs.tex
\newcommand{\cmt}[1]{}

\newcommand{\karen}[1]{\textcolor{red}{{Karen: #1}}}

\newcommand{\visak}[1]{\textcolor{blue}{{Visak: #1}}}

\long\def\ignorethis#1{}

\newcommand{\etal}{{\em{et~al.}\ }}
\newcommand{\eg}{e.g.\ }

\newcommand{\figref}[1]{Figure~\ref{fig:#1}}

\newcommand{\tabref}[1]{Table~\ref{tab:#1}}

\newcommand{\vc}[1]{\ensuremath{\mathbf{#1}}}

\newcommand{\pctab}{\hspace{0.2in}}



%% file: related.tex
Researchers have proposed various motion planning and optimal control algorithms to reduce the damage of humanoid falls. One possible approach is to design a few falling motion sequences for a set of expected scenarios. When a fall is detected, the sequence designed for the most similar falling scenario is executed \cite{Fujiwara2002,Ruiz-del-solar2010}. This approach, albeit simple, can be a practical solution in an environment in which the types of falls can be well anticipated. To handle more dynamic environments, a number of researchers cast the falling problem to an optimization which minimizes the damage of the fall. To accelerate the computation, various simplified models have been proposed to approximate falling motions, such as an inverted pendulum \cite{Fujiwara2006, Fujiwara2007}, a planar robot in the sagittal plane \cite{Wang2012}, a tripod \cite{Yun2014}, and a sequence of inverted pendulums \cite{Ha2015}. In spite of the effort to reduce the computation, most of the optimization-based techniques are still too slow for real-time applications, with the exception of the work done by Goswami \etal \cite{Goswami2014}, who proposed to compute the optimal stepping location to change the falling direction. In contrast, our work takes the approach of policy learning using deep reinforcement learning techniques. Once trained, the policy is capable of handling various situations with real-time computation.

Our work is also built upon recent  advancement in deep reinforcement learning (DRL). Although the network architecture used in this work is not necessarily "deep", we borrow many key ideas from the DRL literature to enable training a large network with $278976$ variables. The ideas of "experience replay" and "target network" from Mnih \etal \cite{Mnih2015} are crucial to the efficiency and stability of our learning process, despite that the original work (DQN) is designed for learning Atari video games from pixels with the assumption that the action space is discrete. Lilicrap \etal \cite{Lillicrap2015} later combined the ideas of DQN and the deterministic policy gradient (DPG) \cite{Silver2014} to learn actor-critic networks for continuous action space and demonstrated that end-to-end (vision perception to actuation) policies for dynamic tasks can be efficiently trained. 

The approach of actor-critic learning has been around for many decades \cite{Sutton1988}. The main idea is to simultaneously learn the state-action value function (the Q-function) and the policy function, such that the intractable
optimization of the Q-function over continuous action space can be avoided. van Hasselt and Wiering introduced CACLA \cite{VanHasselt2007, VanHasselt2012} that represents an actor-critic approach using neural networks. Our work adopts the update scheme for the value function and the policy networks proposed in CACLA. Comparing to the recent work using actor-critic networks \cite{Lillicrap2015,Hausknecht2016}, the main difference of CACLA (and our work as well) lies in the update scheme for the actor networks. That is, CACLA updates the actor by matching the action samples while other methods follow the gradients of the accumulated reward function. Our work is mostly related to the MACE algorithm introduced by Peng \etal  \cite{Peng2016}. We adopt their network architecture and the learning algorithm but for a different purpose: instead of using multiple actor-critic pairs to switch between experts, we exploit this architecture to solve an MDP with a mixture of discrete and continuous action variables.

%% file: results.tex
\section{Experiments}

\begin{figure}
\centering
\includegraphics[width=1.0\columnwidth]{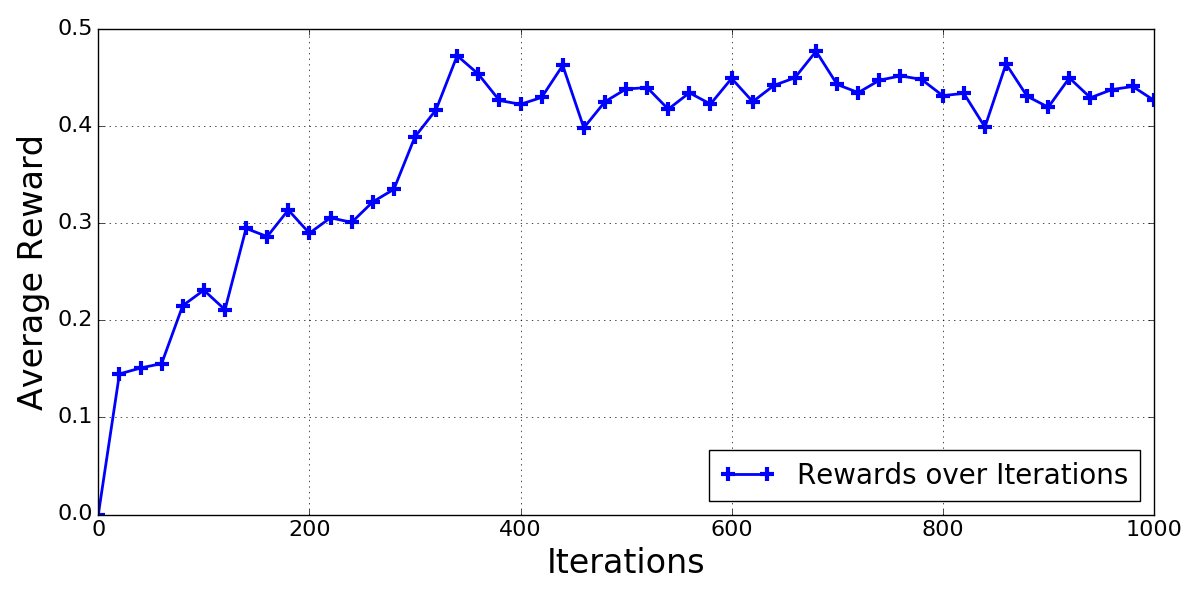}
\caption{The average reward for $10$ test cases.}
\label{fig:reward_iter}
\end{figure}

We validate our policy in both simulation and on physical hardware. The testing platform is a small humanoid, BioloidGP \cite{BioloidGP} with a height of $34.6$~cm, a weight of $1.6$~kg, and 16 actuated degrees of freedom. We also compare the results from our policy against those calculated by the dynamic-programming (DP) based method proposed by Ha and Liu \cite{Ha2015}. Because DP conducts a full search in the action space online, which is 50 to 400 times slower than our policy, in theory DP should produce better solutions than ours. Thus, the goal of the comparison is to show that our policy produces comparable solutions as DP while enjoying the speed gain by two orders of magnitude.


We implement and train the proposed network architecture using PyCaffe \cite{jia2014caffe} on Ubuntu Linux. For simulation, we use a Python binding \cite{PyDart} of an open source physics engine, DART \cite{Dart}.

\subsection{Learning of Abstract-Level Policy}
In our experiment, we construct a network with $8$ pairs of actor-critic to represent $8$ possible contacting body parts: right toe, right heel, left toe, left heel, knees, elbows, hands, and head. During training, we first generate $5000$ tuples from DP to initialize the training buffer. The learning process takes $1000$ iterations, approximately $4$ hours on $6$ cores of 3.3GHz Intel i7 processor. \figref{reward_iter} shows the average reward of $10$ randomly selected test cases over iterations. Once the policy is trained, a single query of the policy network takes approximately $0.8$~milliseconds followed by $25$~milliseconds of the inverse kinematics routine. The total of $25.8$~milliseconds computation time is a drastic improvement from DP which takes $1$ to $10$ seconds of computation time.


\begin{figure}
\centering
\includegraphics[width=1.0\columnwidth]{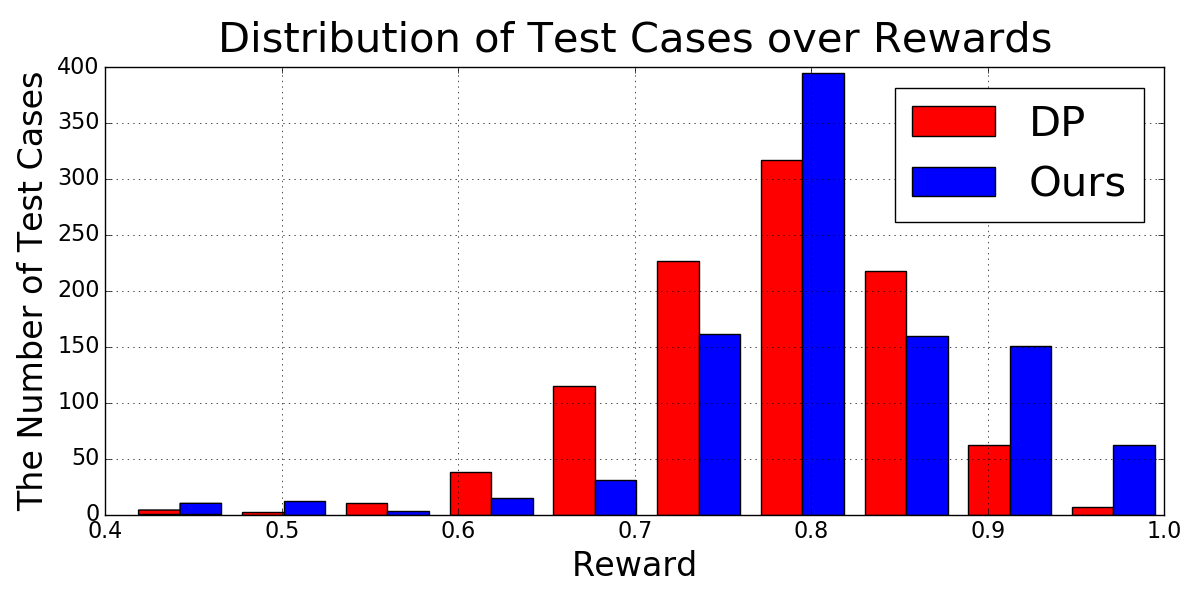}
\caption{The histogram of rewards for the 1000 test cases. Our policy outperforms DP in $65\%$ of the tests.}
\label{fig:reward_distribution}
\end{figure}

\begin{figure*}
\center
\setlength{\tabcolsep}{1pt}
\renewcommand{\arraystretch}{0.5}
\begin{tabular}{c c c c c}
\includegraphics[width=0.195\textwidth]{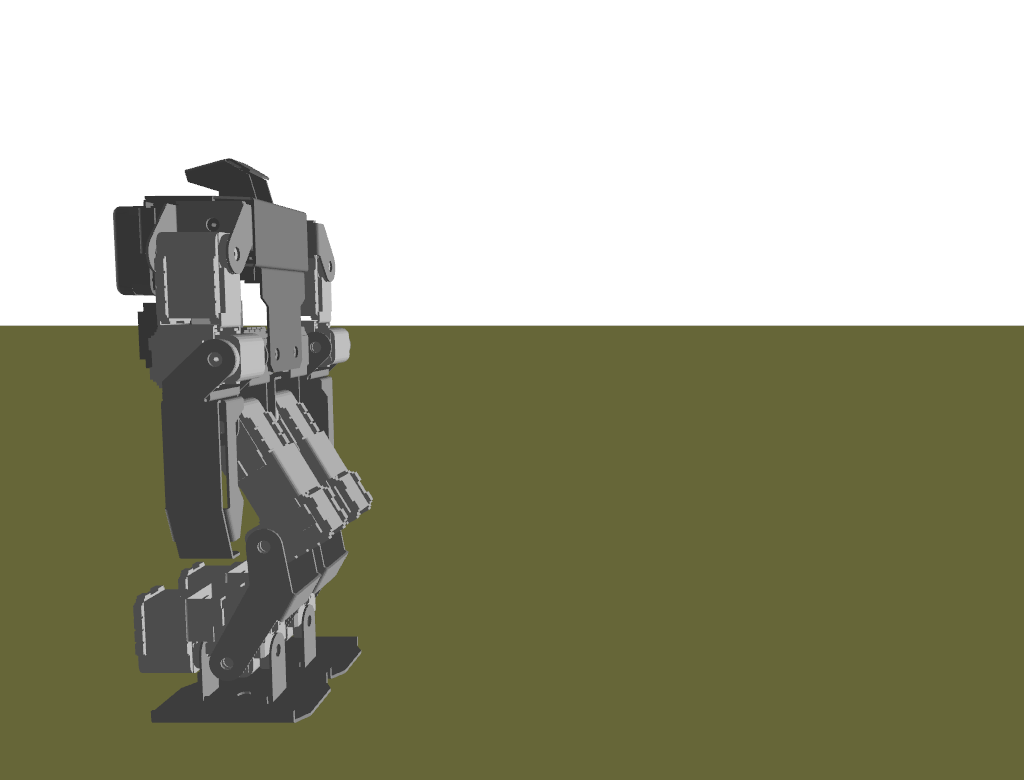}&
\includegraphics[width=0.195\textwidth]{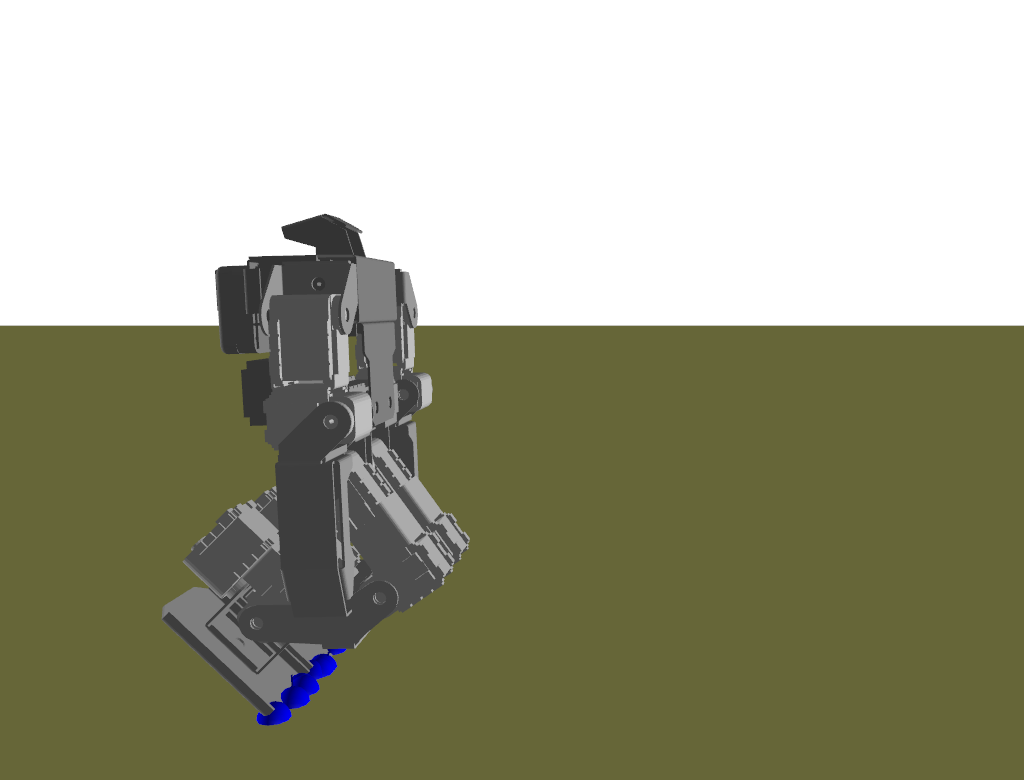}&
\includegraphics[width=0.195\textwidth]{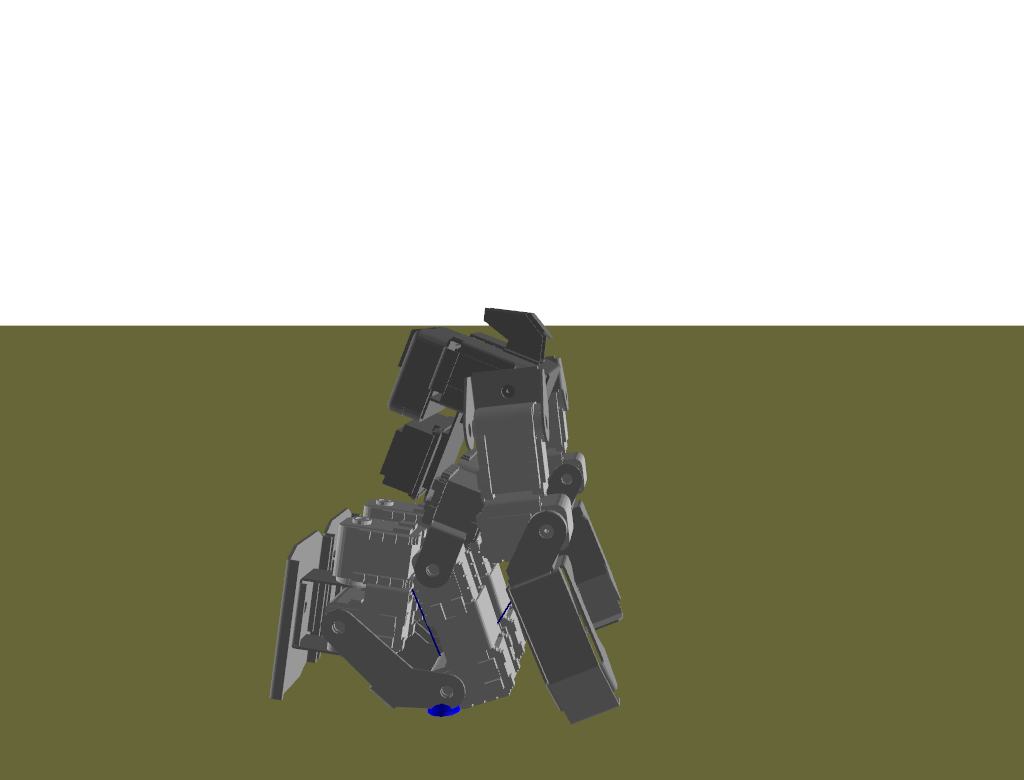}&
\includegraphics[width=0.195\textwidth]{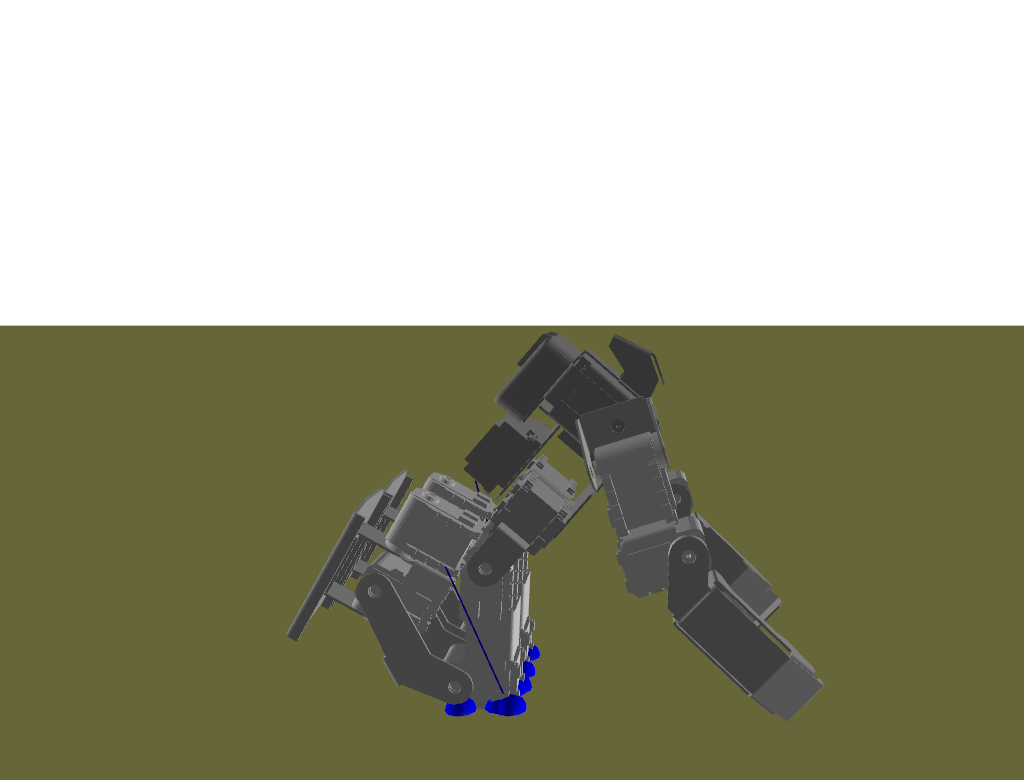}&
\includegraphics[width=0.195\textwidth]{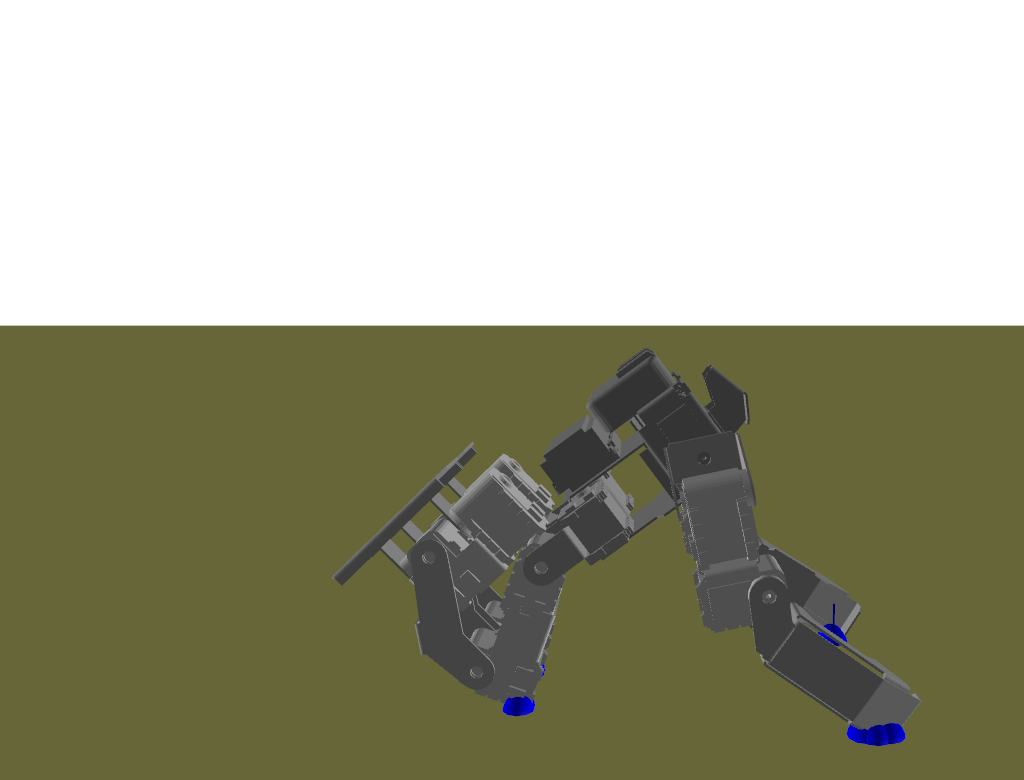} \\
\includegraphics[width=0.195\textwidth]{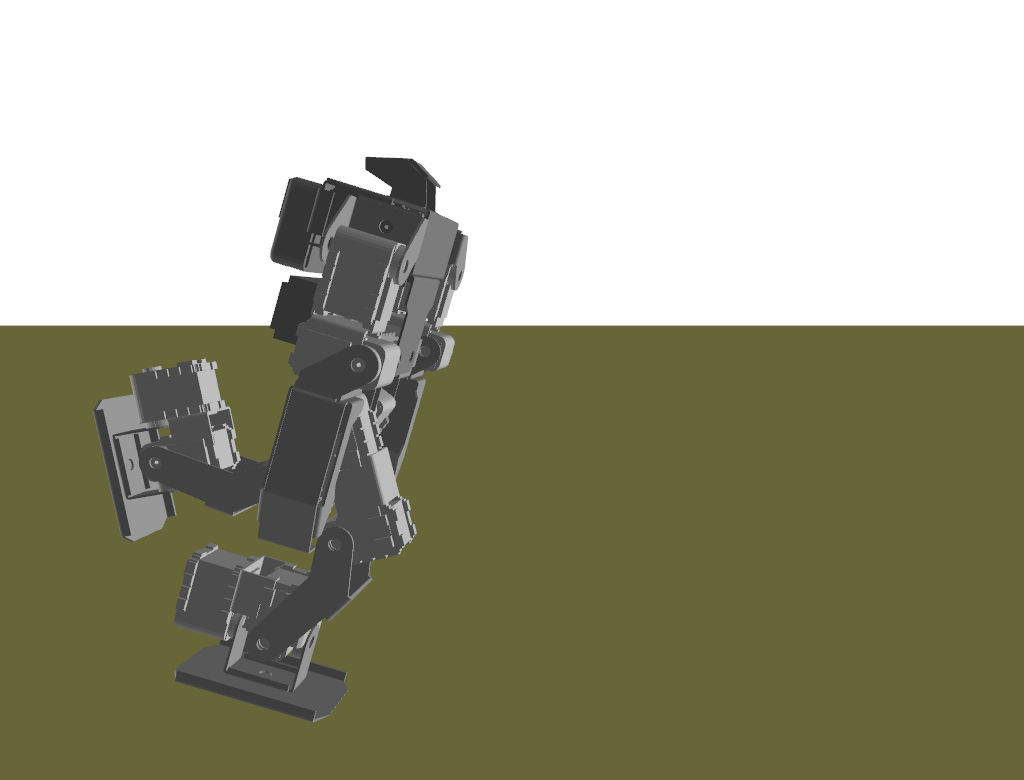}&
\includegraphics[width=0.195\textwidth]{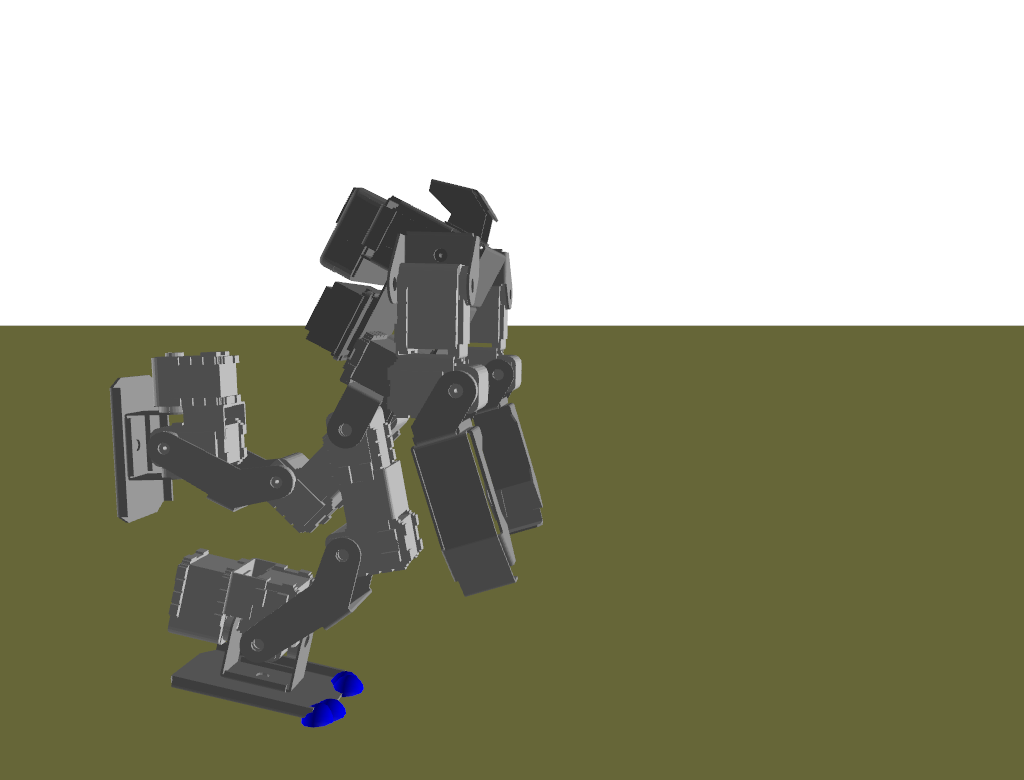}&
\includegraphics[width=0.195\textwidth]{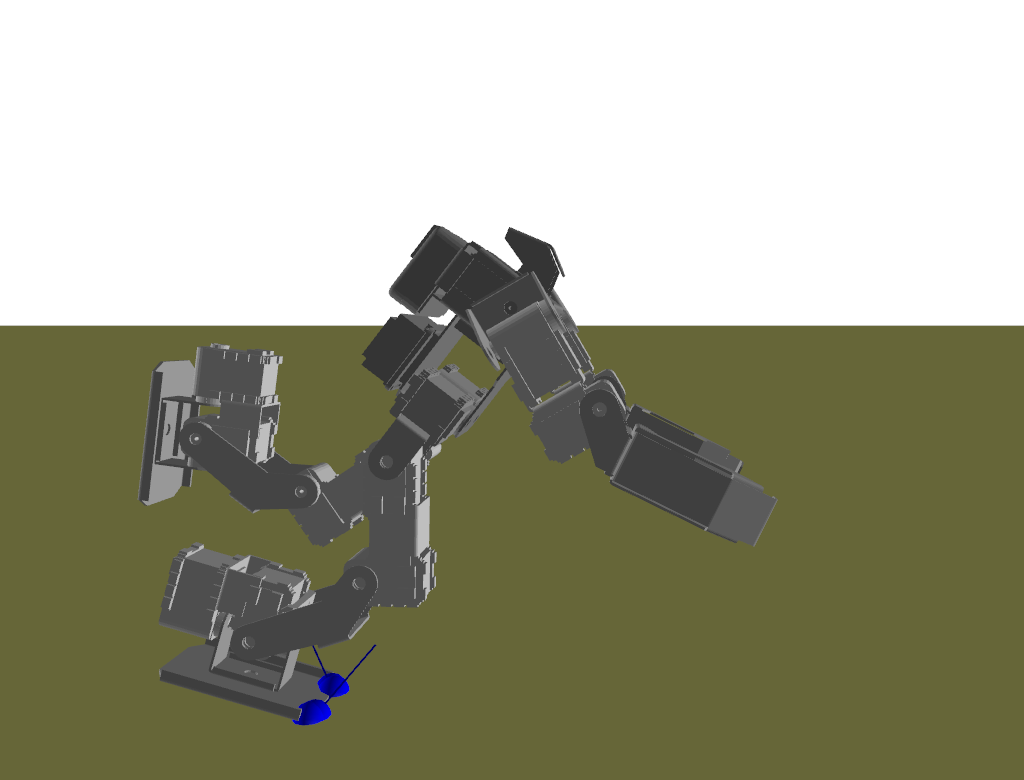}&
\includegraphics[width=0.195\textwidth]{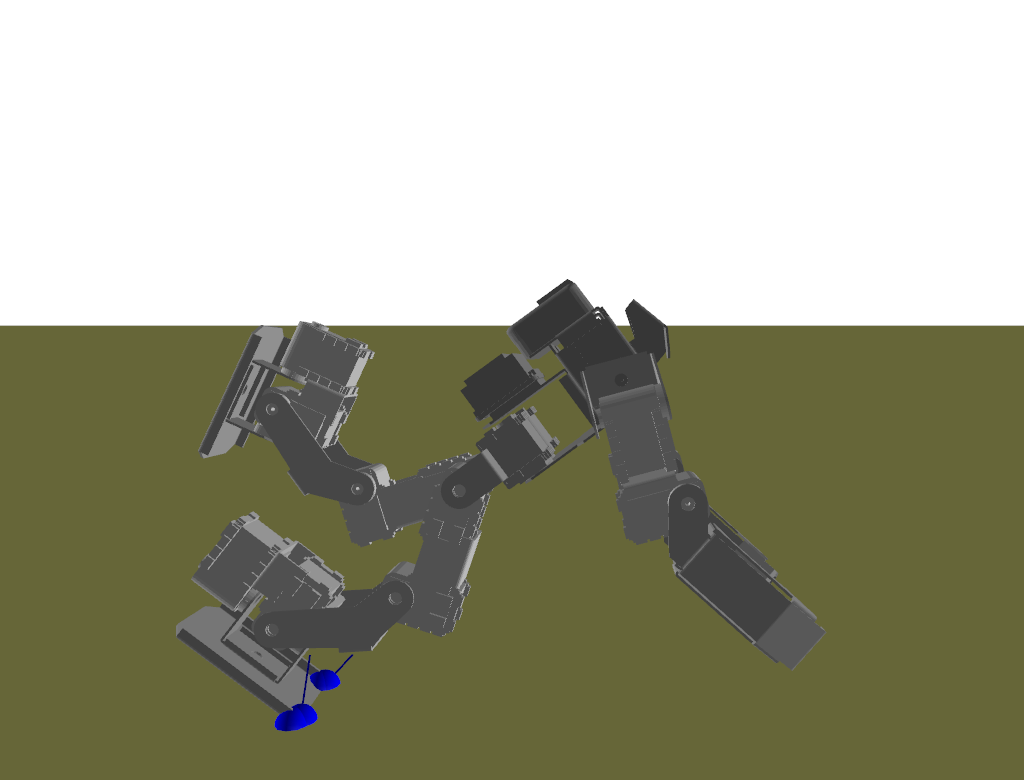}&
\includegraphics[width=0.195\textwidth]{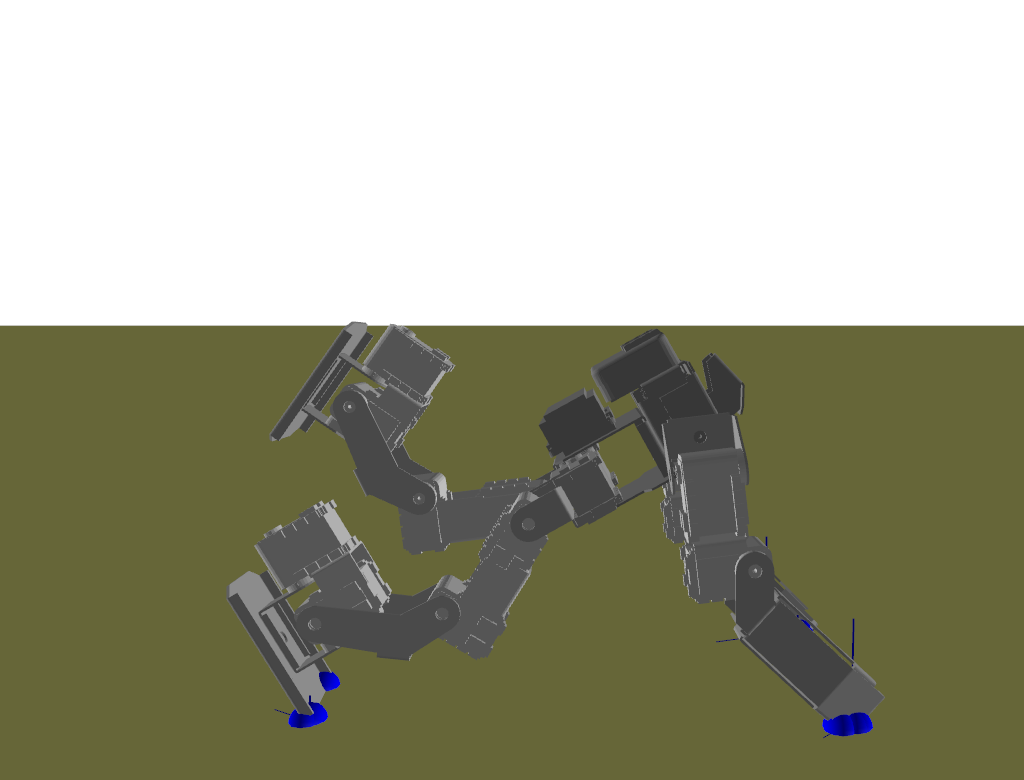} \\
\includegraphics[width=0.195\textwidth]{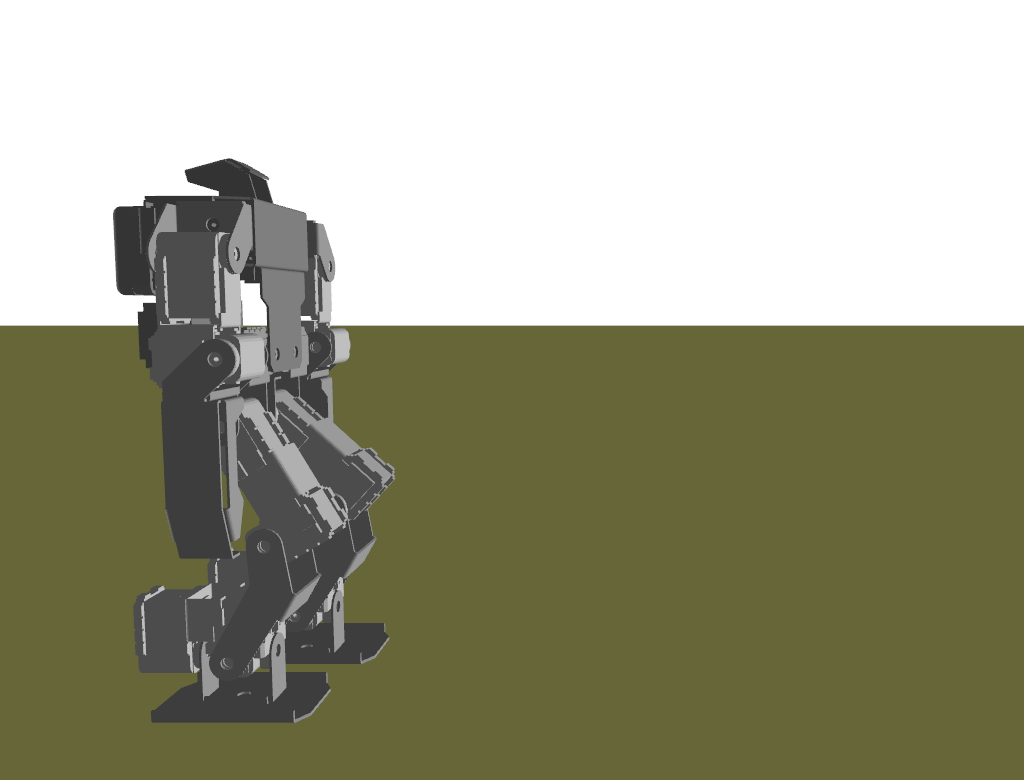}&
\includegraphics[width=0.195\textwidth]{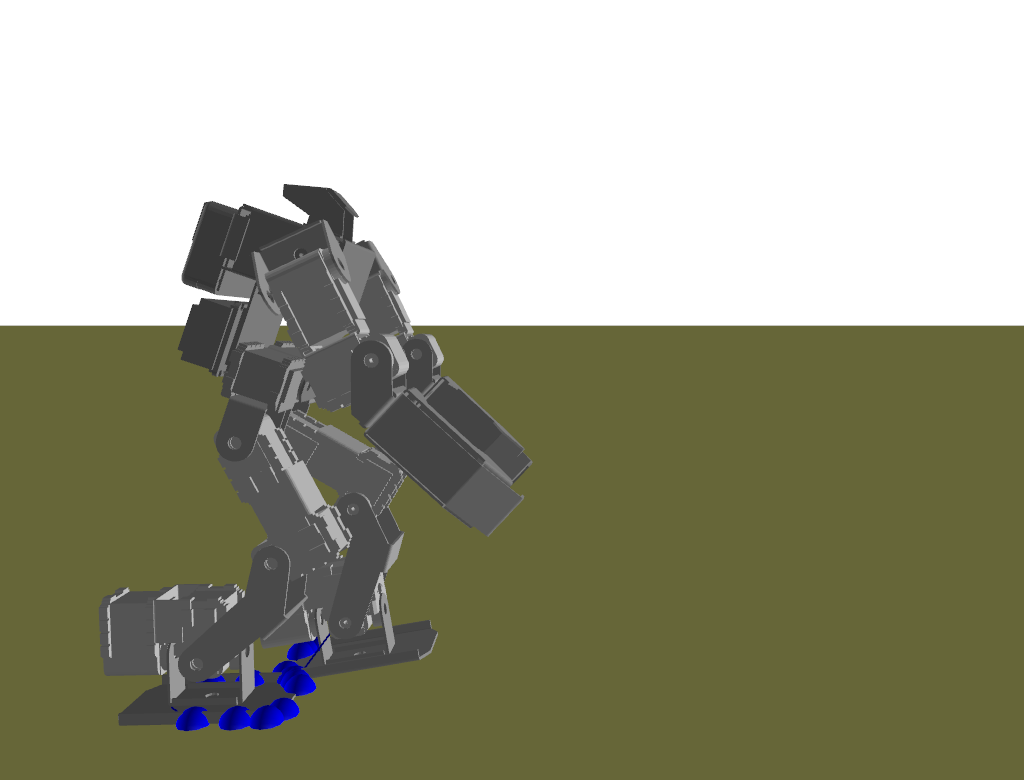}&
\includegraphics[width=0.195\textwidth]{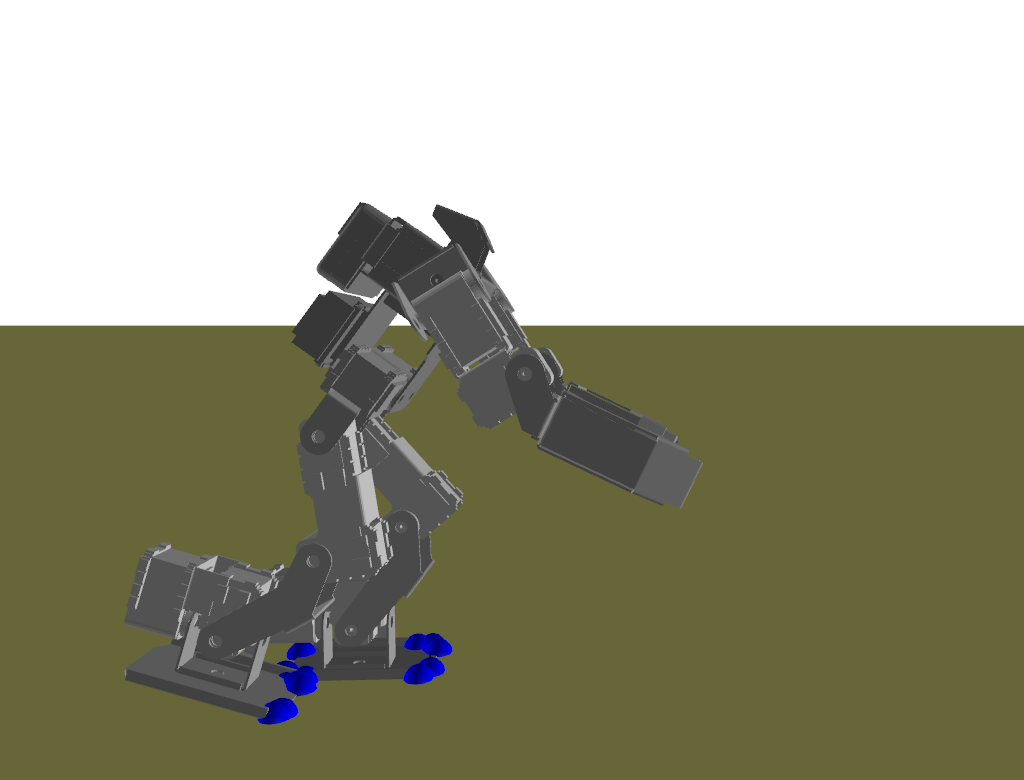}&
\includegraphics[width=0.195\textwidth]{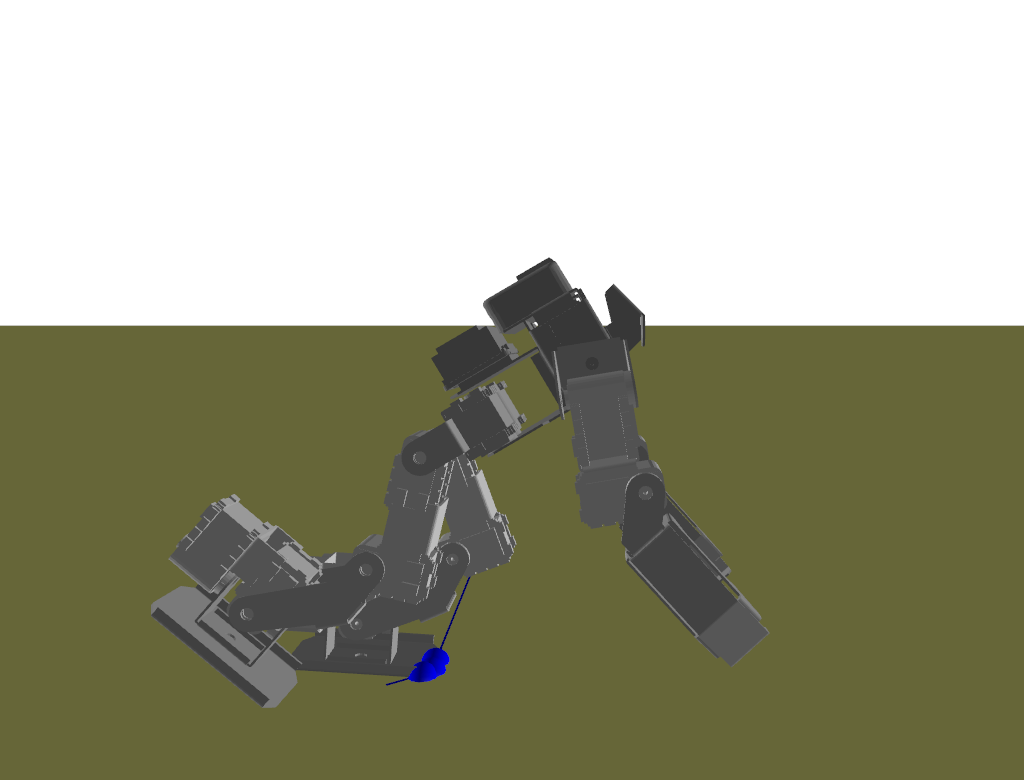}&
\includegraphics[width=0.195\textwidth]{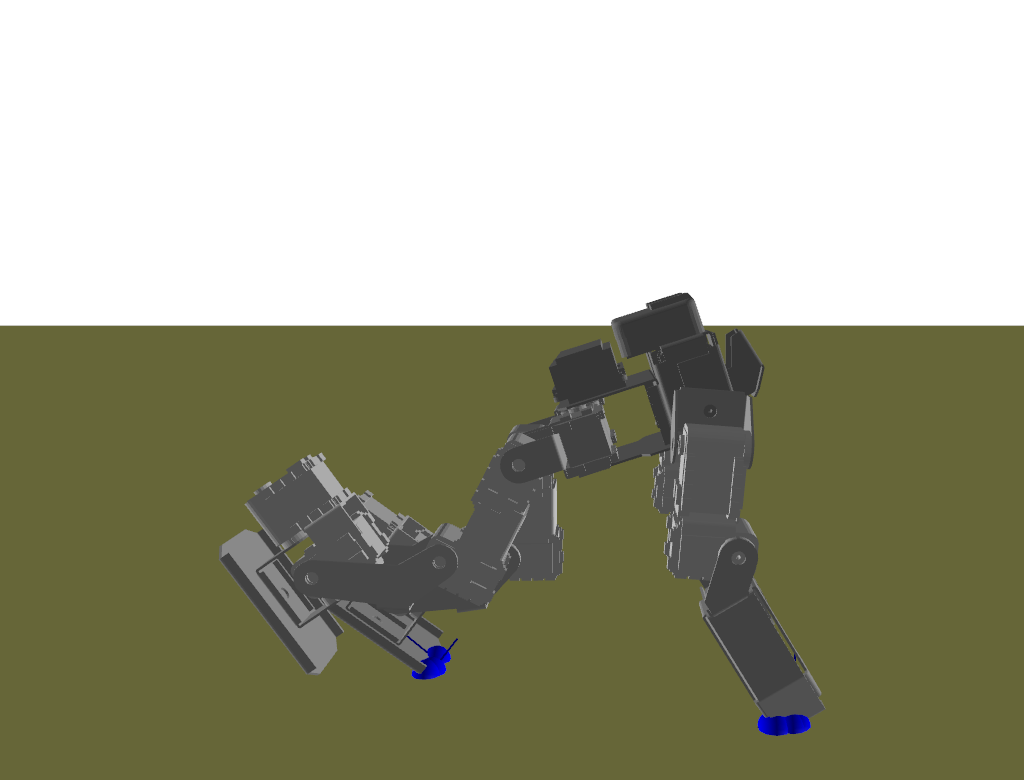} \\
\end{tabular}
\caption{\textbf{Top:} A fall from a two-feet stance due to a $3$~N push (Two-feet). 
         \textbf{Middle:} A fall from an unbalanced stance due to a $5$~N push (Unbalanced). 
         \textbf{Bottom:} A fall from a one-foot stance due to a $6$~N push (One-foot). }
  \label{fig:motions}
\end{figure*}

We compare the rewards of the trained policy with those of DP by running $1000$ test cases starting from randomly sampled initial states. The results are shown in \figref{reward_distribution}, a histogram of rewards for the $1000$ tests computed by our policy and by DP. Our policy achieves not only comparable rewards, it actually outperforms DP in $64$~\% of the test cases. The average reward of our policy is $0.8093$, comparing to $0.7784$ of DP. In terms of impulse, the average of maximum impulse produced by our policy is $0.2540$, which shows a $15$~\% improvement from $0.2997$ produced by DP.

Theoretically, DP, which searches the entire action space for the given initial state, should be the upper bound of the reward our policy can ever achieve. In practice, the discretization of the state and action spaces in DP might result in suboptimal plans. In contrast, our policy exploits the mixture of actor and critic network to optimize continuous action variables without discretization. The more precise continuous optimization often results in more optimal contact location or timing. In some cases, it also results in different contact sequences being selected (45 out of 641 test cases where our policy outperforms DP).




\subsection{Different falling strategies}

\begin{table}
\setlength{\tabcolsep}{3pt}
\renewcommand{\arraystretch}{1.1}
\scriptsize
\center
{
\caption{Different falling strategies.}
\begin{tabular}{| c | c | c |c | c |} 
\hline
 \makecell{\textbf{Initial} \\ \textbf{Pose}} & \textbf{Push} & \textbf{Algorithm} 
& \textbf{Contact Sequence} & \makecell{\textbf{Maximum} \\ \textbf{Impulse}} \\ \hline \hline
&       & Unplanned & Torso ($0.90$) & $0.90$ \\ \cline{3-5}
Two-feet & $3$~N & DP        & Hands ($0.58$) & $0.58$ \\ \cline{3-5}
&       & Ours      & Knees ($0.53$), Hands ($0.20$) & $0.53$ \\ \hline \hline
                    
&       & Unplanned & Torso ($2.50$) & $2.50$ \\ \cline{3-5}
Unbalanced & $5$~N & DP        & Hands ($0.53$) & $0.53$ \\ \cline{3-5}
&       & Ours      & Hands ($0.45$) & $0.45$ \\ \hline \hline
          
&       & Unplanned & Torso ($2.10$) & $2.10$ \\ \cline{3-5}
One-foot& $6$~N & DP         & L-Heel ($0.20$), Hands ($0.45$) & $0.45$ \\ \cline{3-5}
&       & Ours        & L-Heel ($0.10$), Hands ($0.43$) & $0.43$ \\ \hline  
\end{tabular} \label{tab:simulation_results}
}
\end{table}

\begin{figure}
\centering
\includegraphics[width=1.0\columnwidth]{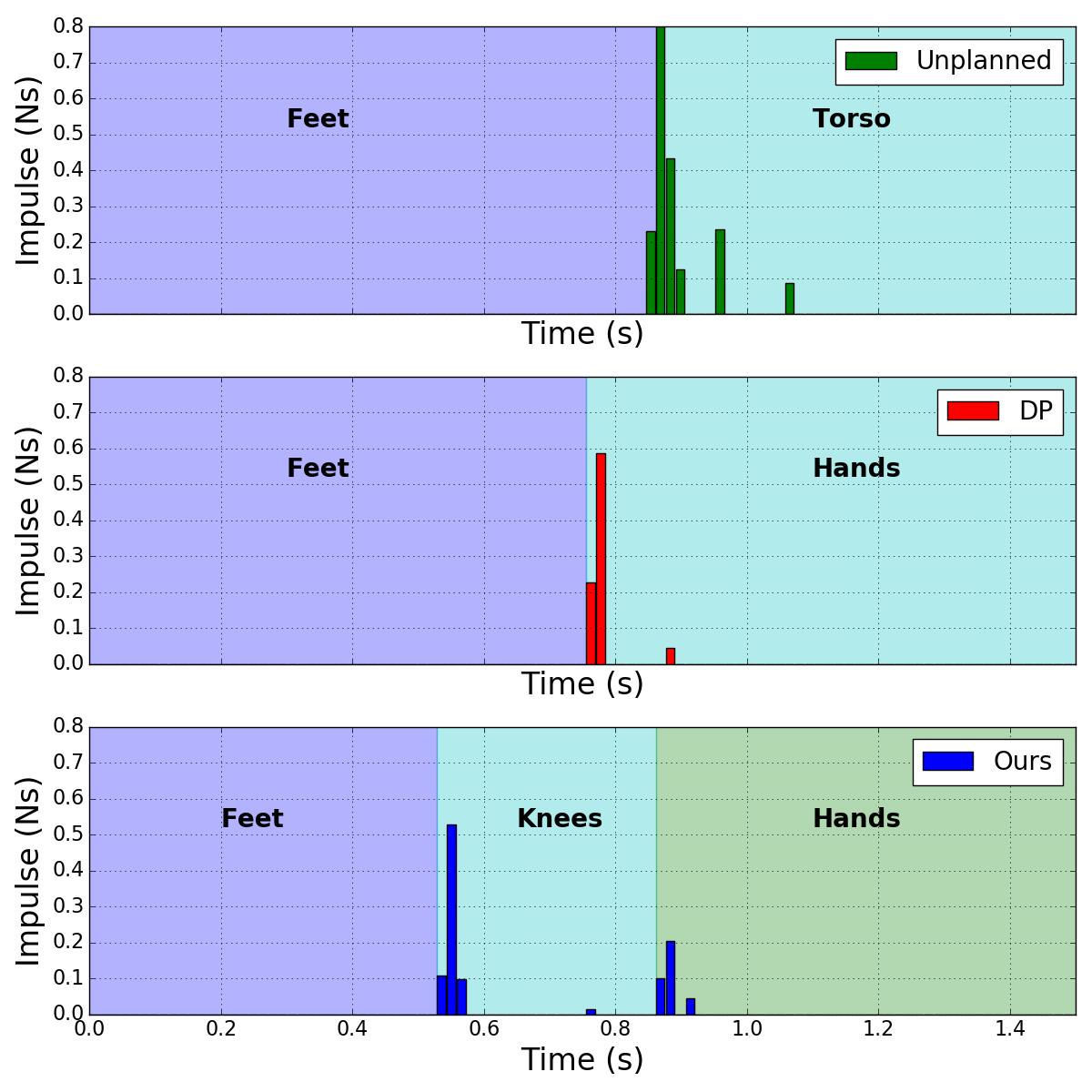}
\caption{Comparison of the impulse profiles among the unplanned motion, the motion planned by DP, and the motion planned by our policy.}
\label{fig:impulse_profile}
\end{figure}

With different initial conditions, various falling strategies emerge as the solution computed by our policy. \tabref{simulation_results} showcases three distinctive falling strategies from the test cases. The table shows the initial pose, the external push, the resulting contact sequence, the impulse due to each contact (the number in the parenthesis), as well as the maximal impulse for each test case. Starting with a two-feet stance, the robot uses knees and hands to stop a fall. If the initial state is leaning forward and unbalanced, the robot directly uses its hands to catch itself. If the robot starts with a one-foot stance, it is easer to use the swing foot followed by the hands to stop a fall. The robot motion sequences can be visualized in Figure \ref{fig:motions} and the supplementary video. For each case, we compare our policy against DP and a naive controller which simply tracks the initial pose (referred as \emph{Unplanned}). Both our policy and DP significantly reduce the maximum impulse comparing to Unplanned. In the cases where our policy outperforms DP, the improvement can be achieved by different contact timing (One-foot case), better target poses (Unbalanced case), or different contact sequences (Two-feet case, \figref{impulse_profile}).

\subsection{Hardware Results}

\begin{figure}
\center
\setlength{\tabcolsep}{1pt}
\renewcommand{\arraystretch}{0.5}
\begin{tabular}{c}
\includegraphics[width=1.0\columnwidth]{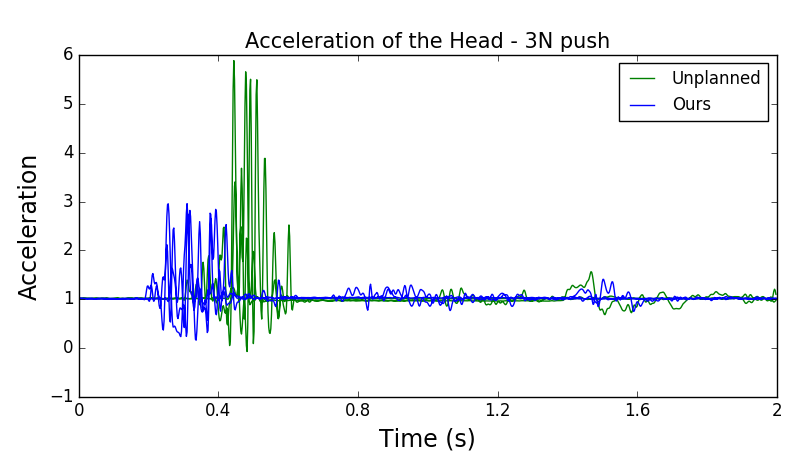} \\
\includegraphics[width=1.0\columnwidth]{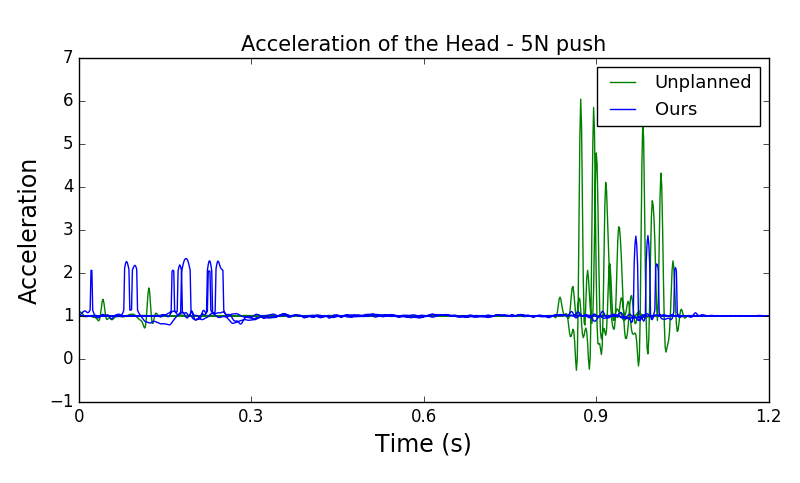} \\
\end{tabular}
\caption{Comparison of measured acceleration between motion computed by our policy and unplanned motion. Three trials for each condition are plotted.
         \textbf{Top:} A fall from a two-feet stance due to a $3$~N push.  
         \textbf{Bottom:} A fall from an one-foot stance due to a $5$~N push. }
  \label{fig:hardware_accel}
\end{figure}

Finally, we compare the falling strategy generated by our policy against the unplanned motion on the hardware of BioloidGP. Due to the lack of on-board sensing capability, BioloidGP cannot take advantage of the feedback aspect of our policy. Nevertheless, we can still use this platform to demonstrate the falling strategy generated by our policy and compare it against an unplanned motion.

We first match the initial pose of the simulated BioloidGP with the real one and push the simulated BioloidGP from the back by $3$~N and $5$~N, assuming that the pushes we applied to the robot by hand are about the same. We then apply our policy on the simulated BioloidGP to obtain a sequence of target poses. In the hardware experiment, we program BioloidGP to track these poses once a fall is detected. During the falls, we measure the acceleration of the head using an external IMU. Figure \ref{fig:hardware_accel} shows the results of two different falls. In the first case, the robot is pushed with a force of 3N and is initialized with both the feet on the ground and an upright position, the robot uses its knees first and then the hands to control the fall. The maximal acceleration from our policy is $2.9$~G while that from an unplanned motion is $5.7$~G, showing a $49$\% of improvement. In the second case, the robot is pushed with a force of 5N starting with one foot on the ground, the falling strategy for this includes using the left-heel first then the hands to control the fall. The maximal acceleration from our policy is $2.3$~G while that from an unplanned motion is $6.4$~G, showing a $64$\% of improvement.

%% file: conclusion.tex
\section{Conclusion}
We proposed a new policy optimization method to learn the appropriate actions for minimizing the damage of a humanoid fall. Unlike most optimal control problems, the action space of our problem consists of both discrete and continuous variables. To address this issue, our algorithm trains $n$ control policies (actors) and the corresponding value functions (critics) in an actor-critic network. Each actor-critic pair is associated with a candidate contacting body part. When the robot establishes a new contact with the ground, the policy corresponding to the highest value function will be executed while the associated body part will be the next contact. As a result of this mixture of actor-critic architecture, we cast the discrete contact planning into the problem of expert selection, while optimizing the policy in continuous space. We show that our algorithm reliably reduces the maximal impulse of a variety of falls. Comparing to the previous work \cite{Ha2015} that employs an expensive dynamic programming method during online execution, our policy can reach better reward and only takes $0.25\%$ to $2\%$ of computation time on average.

One limitation of this work is the assumption that humanoid falls primarily lie on the sagittal plane. This limitation is due to our choice of the simplified model, which reduces computation time but only models planar motions. This assumption can be easily challenged when considering real-world falling scenarios, such as those described in \cite{Yun2014, Goswami2014}. One possible solution to handling more general falls is to employ a more complex model similar to the inertia-loaded inverted pendulum proposed by \cite{Goswami2014}.

Another possible future work direction is to learn control policies directly in the full-body joint space, bypassing the need of an abstract model and the restrictions come with it. This allows us to consider more detailed features of the robot during training, such as full body dynamics or precise collision shapes. Given the increasingly more powerful policy learning algorithms for deep reinforcement learning \cite{schulman2015trust,schulman2015high}, motor skill learning with a large number of variables, as is the case with falling, becomes a feasible option.


%% file: root.bbl
\begin{thebibliography}{10}
\providecommand{\url}[1]{#1}
\csname url@samestyle\endcsname
\providecommand{\newblock}{\relax}
\providecommand{\bibinfo}[2]{#2}
\providecommand{\BIBentrySTDinterwordspacing}{\spaceskip=0pt\relax}
\providecommand{\BIBentryALTinterwordstretchfactor}{4}
\providecommand{\BIBentryALTinterwordspacing}{\spaceskip=\fontdimen2\font plus
\BIBentryALTinterwordstretchfactor\fontdimen3\font minus
  \fontdimen4\font\relax}
\providecommand{\BIBforeignlanguage}[2]{{%
\expandafter\ifx\csname l@#1\endcsname\relax
\typeout{** WARNING: IEEEtran.bst: No hyphenation pattern has been}%
\typeout{** loaded for the language `#1'. Using the pattern for}%
\typeout{** the default language instead.}%
\else
\language=\csname l@#1\endcsname
\fi
#2}}
\providecommand{\BIBdecl}{\relax}
\BIBdecl

\bibitem{Ha2015}
S.~Ha and C.~K. Liu, ``{Multiple Contact Planning for Minimizing Damage of
  Humanoid Falls},'' \emph{IEEE IEEE/RSJ International Conference on
  Intelligent Robots and Systems}, 2015.

\bibitem{VanHasselt2007}
H.~Van~Hasselt and M.~A. Wiering, ``{Reinforcement Learning in Continuous
  Action Spaces},'' no. Adprl, pp. 272--279, 2007.

\bibitem{VanHasselt2012}
H.~Van~Hasselt, ``{Reinforcement Learning in Continuous State and Action
  Spaces},'' \emph{Reinforcement Learning}, pp. 207----251, 2012.

\bibitem{Peng2016}
X.~B. Peng, G.~Berseth, and M.~van~de Panne, ``{Terrain-Adaptive Locomotion
  Skills using Deep Reinforcement Learning},'' \emph{ACM Transactions on
  Graphics}, vol.~35, no.~4, pp. 1--10, 2016.

\bibitem{Fujiwara2002}
K.~Fujiwara, F.~Kanehiro, S.~Kajita, K.~Kaneko, K.~Yokoi, and H.~Hirukawa,
  ``{UKEMI: falling motion control to minimize damage to biped humanoid
  robot},'' \emph{IEEE/RSJ International Conference on Intelligent Robots and
  Systems}, vol.~3, no. October, 2002.

\bibitem{Ruiz-del-solar2010}
J.~Ruiz-del solar, S.~Member, J.~Moya, and I.~Parra-tsunekawa, ``{Fall
  Detection and Management in Biped Humanoid Robots},'' \emph{Management},
  vol.~12, no. April, pp. 3323--3328, 2010.

\bibitem{Fujiwara2006}
K.~Fujiwara, S.~Kajita, K.~Harada, K.~Kaneko, M.~Morisawa, F.~Kanehiro,
  S.~Nakaoka, and H.~Hirukawa, ``{Towards an optimal falling motion for a
  humanoid robot},'' \emph{Proceedings of the 2006 6th IEEE-RAS International
  Conference on Humanoid Robots, HUMANOIDS}, pp. 524--529, 2006.

\bibitem{Fujiwara2007}
------, ``{An Optimal planning of falling motions of a humanoid robot},''
  \emph{IEEE International Conference on Intelligent Robots and Systems}, no.
  Table I, pp. 456--462, 2007.

\bibitem{Wang2012}
\BIBentryALTinterwordspacing
J.~Wang, E.~Whitman, and M.~Stilman, ``{Whole-body trajectory optimization for
  humanoid falling},'' \emph{{\ldots} Control Conference (ACC), {\ldots}}, pp.
  4837--4842, 2012. [Online]. Available:
  \url{http://ieeexplore.ieee.org/xpls/abs{\_}all.jsp?arnumber=6315177}
\BIBentrySTDinterwordspacing

\bibitem{Yun2014}
S.~K. Yun and A.~Goswami, ``{Tripod fall: Concept and experiments of a novel
  approach to humanoid robot fall damage reduction},'' \emph{Proceedings - IEEE
  International Conference on Robotics and Automation}, pp. 2799--2805, 2014.

\bibitem{Goswami2014}
A.~Goswami, S.-k. Yun, U.~Nagarajan, S.-H. Lee, K.~Yin, and S.~Kalyanakrishnan,
  ``{Direction-changing fall control of humanoid robots: theory and
  experiments},'' \emph{Autonomous Robots}, vol.~36, no.~3, pp. 199--223, jul
  2014.

\bibitem{Mnih2015}
V.~Mnih, K.~Kavukcuoglu, D.~Silver, A.~a. Rusu, J.~Veness, M.~G. Bellemare,
  A.~Graves, M.~Riedmiller, A.~K. Fidjeland, G.~Ostrovski, S.~Petersen,
  C.~Beattie, A.~Sadik, I.~Antonoglou, H.~King, D.~Kumaran, D.~Wierstra,
  S.~Legg, and D.~Hassabis, ``{Human-level control through deep reinforcement
  learning},'' \emph{Nature}, vol. 518, no. 7540, pp. 529--533, 2015.

\bibitem{Lillicrap2015}
T.~P. Lillicrap, J.~J. Hunt, A.~Pritzel, N.~Heess, T.~Erez, Y.~Tassa,
  D.~Silver, and D.~Wierstra, ``{Continuous control with deep reinforcement
  learning},'' \emph{arXiv preprint arXiv:1509.02971}, pp. 1--14, 2015.

\bibitem{Silver2014}
D.~Silver, G.~Lever, N.~Heess, T.~Degris, D.~Wierstra, and M.~Riedmiller,
  ``{Deterministic Policy Gradient Algorithms},'' \emph{Proceedings of the 31st
  International Conference on Machine Learning}, 2014.

\bibitem{Sutton1988}
R.~S. Sutton and A.~G. Barto, \emph{{Reinforcement learning : an
  introduction}}, 1988.

\bibitem{Hausknecht2016}
\BIBentryALTinterwordspacing
M.~Hausknecht and P.~Stone, ``{Deep Reinforcement Learning in Parameterized
  Action Space},'' \emph{arXiv}, pp. 1--12, 2016. [Online]. Available:
  \url{http://arxiv.org/abs/1511.04143}
\BIBentrySTDinterwordspacing

\bibitem{BioloidGP}
\emph{BioloidGP, http://en.robotis.com/}.

\bibitem{jia2014caffe}
Y.~Jia, E.~Shelhamer, J.~Donahue, S.~Karayev, J.~Long, R.~Girshick,
  S.~Guadarrama, and T.~Darrell, ``Caffe: Convolutional architecture for fast
  feature embedding,'' \emph{arXiv preprint arXiv:1408.5093}, 2014.

\bibitem{PyDart}
PyDart, \emph{A Python Binding of Dynamic Animation and Robotics Toolkit,
  http://pydart2.readthedocs.io}.

\bibitem{Dart}
DART, \emph{Dynamic Animation and Robotics Toolkit, http://dartsim.github.io/}.

\bibitem{schulman2015trust}
J.~Schulman, S.~Levine, P.~Moritz, M.~I. Jordan, and P.~Abbeel, ``Trust region
  policy optimization,'' \emph{CoRR, abs/1502.05477}, 2015.

\bibitem{schulman2015high}
J.~Schulman, P.~Moritz, S.~Levine, M.~Jordan, and P.~Abbeel, ``High-dimensional
  continuous control using generalized advantage estimation,'' \emph{arXiv
  preprint arXiv:1506.02438}, 2015.

\end{thebibliography}
